\PassOptionsToPackage{table}{xcolor}

\documentclass[letterpaper, 10 pt, conference]{class/ieeeconf}  

\IEEEoverridecommandlockouts                              
\usepackage{anyfontsize}
\usepackage{ stmaryrd }
\usepackage{amsmath,amsfonts,bm}
\usepackage{cuted}
\usepackage[ruled,vlined,linesnumbered]{algorithm2e}
\usepackage{graphicx}
\usepackage{tabularx} 

\usepackage{amssymb}
\usepackage{amsthm}
\usepackage{multirow}

\usepackage{floatrow}
\usepackage{color}
\usepackage[bb=boondox]{mathalfa}
\usepackage{dsfont}
\usepackage{upgreek}
\usepackage{mathrsfs}
\usepackage{url}
\usepackage{accents}

\usepackage{pifont}
\newcommand{\xmark}{\ding{55}}%

\def\eqref#1{equation~\ref{#1}}

\def\1{\bm{1}}

\DeclareMathAlphabet{\mathsfit}{\encodingdefault}{\sfdefault}{m}{sl}
\SetMathAlphabet{\mathsfit}{bold}{\encodingdefault}{\sfdefault}{bx}{n}

\usepackage[table,xcdraw]{xcolor}
\usepackage{cite}
\usepackage{latexsym}

\usepackage{flushend}
\usepackage{lipsum}

\usepackage{cite}
\usepackage{relsize}
\usepackage{multirow}
\usepackage{afterpage}
\usepackage{ifthen}
\usepackage{graphicx}
\usepackage{algorithm2e}
\usepackage{subfigure}
\usepackage{epstopdf}
\usepackage{stackengine}
\usepackage{gensymb}
\usepackage{booktabs}
\usepackage{makecell}
\usepackage{hhline}
\usepackage{ upgreek }
\usepackage{mathtools}
\usepackage{caption}

\usepackage{amsthm}

\makeatletter
\let\NAT@parse\undefined
\makeatother
\usepackage[bookmarks=false, linkcolor=blue, urlcolor=blue, citecolor=blue]{hyperref} 

\hypersetup{
    colorlinks=true,
    linkcolor=red,
    filecolor=magenta,      
    urlcolor=blue,
    pdfstartview={FitH},
    citecolor =blue
    }

\title{\LARGE \bf AeroScene: Progressive Scene Synthesis for Aerial Robotics 
}

\author{Nghia Vu$^{\dagger, 2}$, Tuong Do$^{\dagger,1,2,3}$, Dzung Tran$^{4}$, Binh X. Nguyen$^{2}$, Hoan Nguyen$^{5}$, Erman Tjiputra$^{2}$, \\ Quang D. Tran$^{1,2}$, Hai-Nguyen Nguyen$^{4}$, Anh Nguyen$^{1}$
}

\begin{document}
\twocolumn[{%
\renewcommand\twocolumn[1][]{#1}%
\maketitle

\begin{center}
    \centering
    \vspace{-5ex}
    \includegraphics[width=\textwidth]{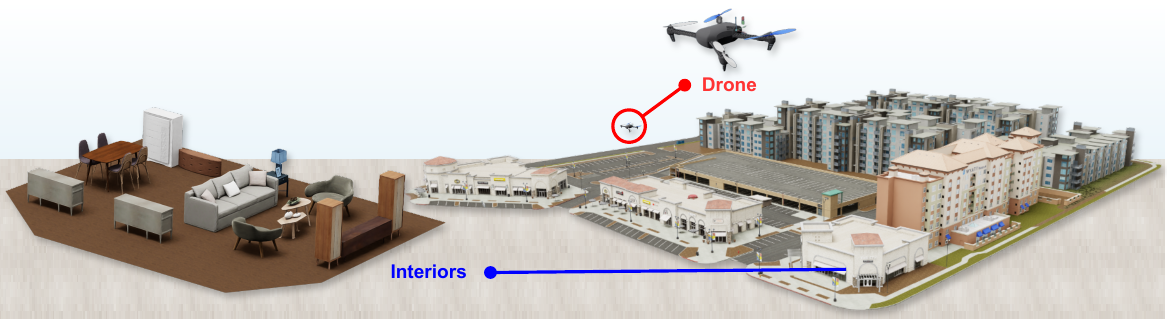}
    \captionof{figure}{We introduce AeroScene, a progressive scene synthesis method and dataset for aerial robotics.
    }
    \label{fig:intro}
\end{center}%
}]






\begin{abstract}
Generative models have shown substantial impact across multiple domains, their potential for scene synthesis remains underexplored in robotics. This gap is more evident in drone simulators, where simulation environments still rely heavily on manual efforts, which are time-consuming to create and difficult to scale. In this work, we introduce AeroScene, a hierarchical diffusion model for progressive 3D scene synthesis. Our approach leverages hierarchy-aware tokenization and multi-branch feature extraction to reason across both global layouts and local details, ensuring physical plausibility and semantic consistency. This makes AeroScene particularly suited for generating realistic scenes for aerial robotics tasks such as navigation, landing, and perching. We demonstrate its effectiveness through extensive experiments on our newly collected dataset and a public benchmark, showing that AeroScene significantly outperforms prior methods. Furthermore, we use AeroScene to generate a large-scale dataset of over 1,000 physics-ready, high fidelity 3D scenes that can be directly integrated into NVIDIA Isaac Sim. Finally, we illustrate the utility of these generated environments on downstream drone navigation tasks. Our code and dataset are publicly available at \href{https://aioz-ai.github.io/AeroScene/}{aioz-ai.github.io/AeroScene/}
\end{abstract}
\renewcommand{\thefootnote}{}
\setcounter{footnote}{0}
\footnotetext{
$^{\dagger}$ Equal contribution

$^{1}$ University of Liverpool, UK

$^{2}$ AIOZ Ltd., Singapore

$^{3}$ National Tsing Hua University, Taiwan

$^{4}$ RMIT University, Vietnam Campus

$^{5}$ University of Information Technology, VNUHCM, Vietnam.
}
\section{Introduction}

Drones are increasingly applied in delivery, inspection, and surveillance, requiring them to navigate and operate within complex 3D environments~\cite{ sandikci2025autonomous, cascarano2021design}. Synthesizing realistic scenes for such applications necessitates hierarchical layout generation, where coarse-scale structures (e.g., rooms, terrains, building layouts) establish navigable flight corridors, while fine-scale details (e.g., obstacles, landing areas) ensure task-specific feasibility. However, existing scene creation methods for drone simulators are designed primarily by humans, making them challenging to scale, and often fail to accommodate both physical and fidelity requirements such as unobstructed aerial navigation~\cite{fan2022aerial}, accessible interaction areas (e.g., landing pads, inspection points)~\cite{liu2024aira, vu2026AffordMatcher}, and coherent indoor-outdoor transitions~\cite{pluckter2018precision}.


While several drone simulators have been developed to support research in navigation and control~\cite{furrer2016rotors, shah2017airsim}, most remain limited in providing diverse and realistic environments for evaluating aerial tasks. Many simulators focus on accurate physics and sensor fidelity, yet often rely on static, handcrafted environments, hindering scalability, diversity, and the realism necessary for advanced testing~\cite{nikolaiev2024comparative}. Moreover, interaction areas critical to aerial robotics, such as landing zones, cluttered corridors, and inspection surfaces, are frequently simplified or entirely absent, limiting full task realism~\cite{sabet2022scalable}. As a result, existing platforms offer limited support for benchmarking higher-level autonomy, where navigation, task execution, and environment understanding must be jointly evaluated in realistic, hierarchical settings~\cite{dimmig2023survey}.


In this paper, we propose \textbf{AeroScene}, a hierarchical diffusion-based framework for 3D scene generation tailored to drone tasks. Our approach operates across scales: coarse-scale synthesis generates high-level structures that preserve airspace and navigability, while fine-scale synthesis refines object placement and specifies drone-interaction areas for task execution. AeroScene includes Cross-scale Progressive Attention, which explicitly models dependencies across scales, ensuring that fine-scale details remain consistent with coarse-scale spatial structures. In addition, we design task-aware guidance functions that encourage collision-free plausibility, maintain semantic correlations in hierarchical orders, and handle relationships between indoor and outdoor objects, thereby aligning generated layouts with real-world aerial operation requirements. To support downstream tasks, scenes created by our method are directly embedded into NVIDIA Isaac Sim~\cite{makoviychuk2021isaac} for physics-ready simulation.



Our main contributions are as follows:
\vspace{-1ex}
\begin{itemize}
    \item We introduce a new framework that generates realistic and high-fidelity scenes for aerial robotics. 
    \item We contribute a large-scale dataset with more than 1000 scenes and embed them into Isaac Sim to serve as a benchmark for drone-related tasks. 
\end{itemize}

\section{Related Works}

\textbf{Drone Simulators.} Numerous simulators have been developed to support aerial robotics research, with varying emphasis on physics fidelity, sensor modeling, and environmental complexity~\cite{furrer2016rotors, shah2017airsim}. While these platforms provide valuable testbeds for navigation and perception, most rely on static or handcrafted environments, limiting their ability to represent diverse and scalable 3D scenes. NVIDIA Isaac Sim~\cite{makoviychuk2021isaac} offers a modern foundation with high-quality rendering, physics, and integration with learning frameworks, making it well-suited as a base platform. However, existing simulators focus primarily on physics and sensing rather than adaptive scene synthesis. To highlight these differences, Table~\ref{tab:simulator_comparison} summarizes key features of widely used drone simulators. Our work builds AeroScene to generate large-scale, physics-ready high fidelity scenes for drone-related tasks that can be embedded directly into Issac Sim.

\begin{table}[!ht]
\centering
\caption{Comparison of drone simulators.}
\label{tab:simulator_comparison}
\resizebox{\linewidth}{!}{
\setlength{\tabcolsep}{0.18 em} 
{
\renewcommand{\arraystretch}{1.2}
\begin{tabular}{lccccccc}
\toprule
\textbf{Simulator} & \textbf{Physics} & \textbf{Diversity} & \textbf{Scalability} & \textbf{Indoor} & \textbf{Outdoor} & \textbf{Multi-scale} \\
\midrule
\rowcolor[HTML]{EFEFEF}RotorS~\cite{furrer2016rotors}        & \checkmark & Low    & Limited & \checkmark & \xmark & \xmark \\
AirSim~\cite{shah2017airsim}          & \checkmark & Medium & Limited & \checkmark & \checkmark & \xmark \\
\rowcolor[HTML]{EFEFEF}OmniDrones~\cite{xu2024omnidrones}           & \checkmark & Medium & High     & \checkmark & Partial    & \xmark \\
QuadSwarm~\cite{huang2023quadswarm}          & \checkmark & Low & High    & \checkmark & \xmark     & \xmark \\
\rowcolor[HTML]{EFEFEF}VisFly~\cite{li2024visfly}                   & \checkmark & High   & Medium & Partial & \checkmark & \xmark \\
IAP~\cite{du2025highfidelity}  & \checkmark & High & Medium  & \checkmark & Partial & \xmark \\
\midrule
\rowcolor[HTML]{EFEFEF}\textbf{AeroScene (Ours)}             & \checkmark & \textbf{High} & \textbf{High} & \checkmark & \checkmark & \checkmark \\
\bottomrule
\end{tabular}
}
}
\end{table}

\textbf{Scene Synthesis.} Scene synthesis aims to generate structured 3D layouts of objects in indoor and outdoor environments. Early approaches relied on rule-based or probabilistic grammars~\cite{fu2017automatic} and heuristic priors~\cite{yeh2012synthesizing}, but lacked scalability. Deep generative models improved plausibility, with GAN- and autoregressive frameworks producing realistic layouts \cite{zhang2020learning, lin2023infinicity, xie2024citydreamer, paschalidou2021atiss}, though they often struggle to balance global structure and local detail. Diffusion-based methods offer greater stability and diversity \cite{hoogeboom2022equivariant, tang2024diffuscene, vuong2023language,bokhovkin2025scenefactor}, but typically treat layouts as flat sets, limiting cross-scale reasoning. We tackle this problem with a hierarchical-scale modeling approach, which routes scene elements into coarse and fine branches, fusing them through alternating cross-scale attention. This explicitly propagates global layout context while refining local details, a design particularly suited to drone tasks, where both macro-structure (e.g., roads, buildings) and fine geometry (e.g., vehicles, furniture) are critical for perception, planning, and simulation fidelity \cite{shah2018airsim,madaan2020airsim, wang2018cooperative}.

\textbf{3D Layout Representations.} 
3D scenes have been represented using voxel grids~\cite{maturana2015voxnet,wu20153d, ren2024xcube}, meshes~\cite{tang2024diffuscene, lin2024instructscene, lee2025nuiscene}, point clouds~\cite{qi2017pointnet,qi2017pointnetplusplus}, and object-centric layouts~\cite{fu20213d, tang2024diffuscene, yang2024physcene,yamazaki2024open, deitke2022️}. While voxels and meshes capture high-resolution geometry, they are computationally costly and less interpretable for task-level reasoning. Object-centric layouts abstract scenes into discrete entities with positions, orientations, scales, and semantic labels, providing a compact and interpretable structure suited for simulation and planning~\cite{lee2025dynscene, wang2024architect}. Our method builds on this line and routes objects into coarse-to-fine representations for drone-related tasks.

\textbf{Guided Diffusion Models.}
Guided diffusion has become a popular generative model for task-specific objectives. Classifier guidance~\cite{dhariwal2021diffusion,nguyen2024lightweight}, classifier-free guidance~\cite{ho2022classifierfree}, and score distillation techniques~\cite{poole2023dreamfusion} have enabled control over semantics or conditions. Training-free approaches such as energy-based guidance~\cite{meng2023sdedit}, constraint-driven sampling~\cite{liu2022compositional}, and physical priors~\cite{jiang2023motion} have extended diffusion models to respect external objectives. Prior works in 3D synthesis often adapt generic guidance strategies, such as collision penalties or semantic constraints~\cite{le2023controllable,jain2022zero123,nguyen2024language, ni2024phyrecon}, but typically treat them as auxiliary heuristics rather than deeply integrated objectives. In contrast, we directly incorporate task-specific objectives into the hierarchical scene synthesis process.



\section{Methodology}
\begin{figure*}[!h]
  \centering
  \includegraphics[width=0.85\linewidth]{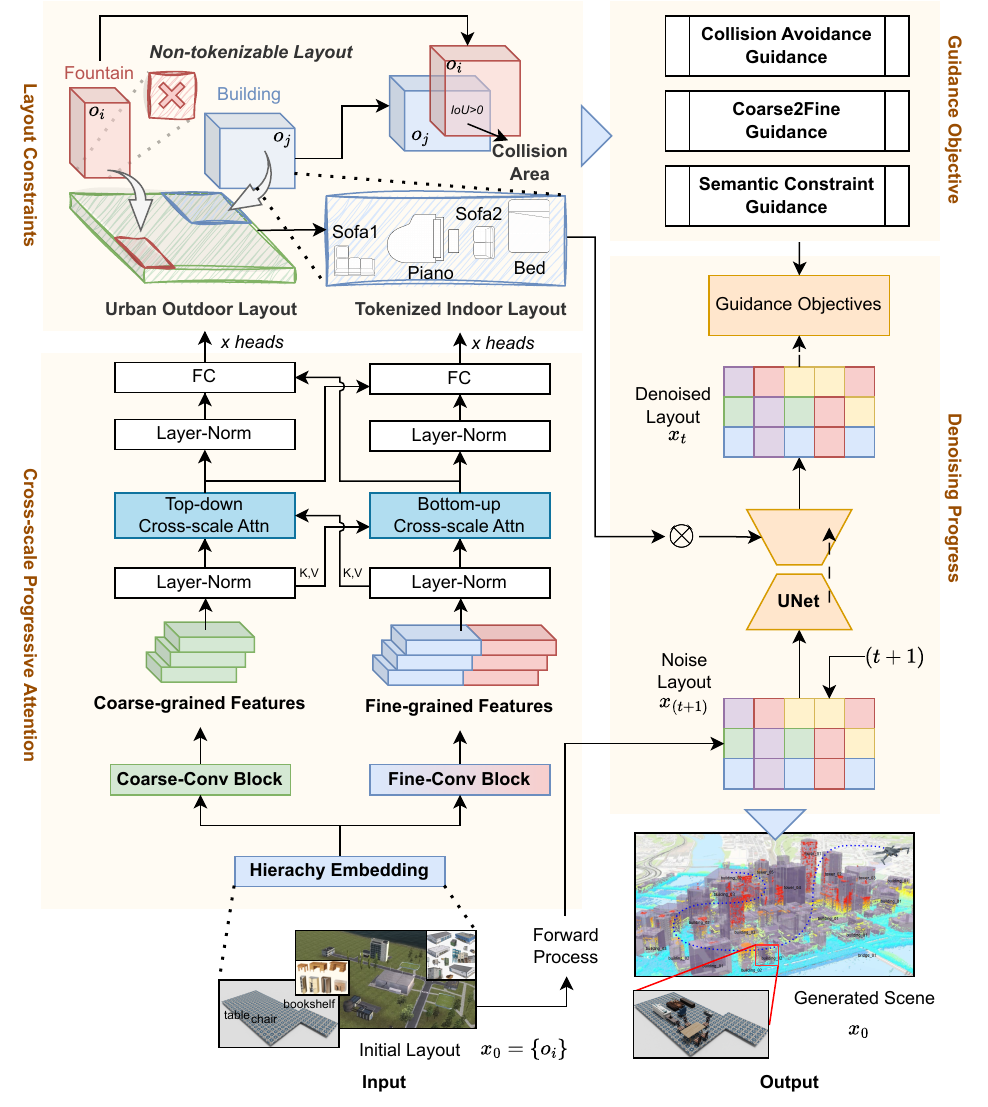}
 \caption{An overview of our AeroScene method. 
 }
 \label{fig:overview}
\end{figure*}

\subsection{Diffusion Process}
We adopt a denoising diffusion probabilistic model (DDPM)~\cite{ho2020denoising} to learn the distribution of plausible scene layouts.
Let $\alpha_t = 1 - \beta_t$ and $\bar{\alpha}_t = \prod_{s=1}^t \alpha_s$ for a fixed noise schedule $\{\beta_t\}_{t=1}^T$. The forward process gradually adds Gaussian noise to a clean layout $x_0$:
\begin{equation}
q(x_t | x_{t-1}) = \mathcal{N}(x_t; \sqrt{\alpha_t}\, x_{t-1}, \beta_t I),
\end{equation}

The reverse process removes noise step-by-step: 
\begin{equation}
p_\theta(x_{t-1} | x_t) = \mathcal{N}(x_{t-1}; \mu_\theta(x_t, t), \Sigma_\theta(x_t,t)),
\end{equation}
where $\mu_\theta$ is predicted by a hierarchy-aware pipeline every denoising timestep. Concretely, at each timestep $t$, we apply the Initial Layout and Hierarchy Embedding (Sec.~\ref{sec:hierarchy_embedding}) to convert $x_t$ into hierarchy tokens, then extract coarse and fine features (Sec.~\ref{sec:feature_extraction}), fuse them with the Cross-scale Progressive Attention (Sec.~\ref{sec:hst}), and finally condition the denoising UNet on the fused features. Guidance objectives (Sec.~\ref{sec:guidance}) are applied to adjust $\mu_\theta$ before sampling $x_{t-1}$. This design ensures the denoiser performs hierarchical reasoning at each diffusion step, matching the iterative sampling loop. Fig.~\ref{fig:overview} shows the details of our method.

\subsection{Initial Layout and Hierarchy Embedding}
\label{sec:hierarchy_embedding}
At each diffusion step, the noisy layout $x_t$ is represented as
\begin{equation}
x_t = \{ o_i \mid o_i = (\mathbf{p}_i, \mathbf{q}_i, \mathbf{s}_i, c_i) \}_{i=1}^{N_o},
\end{equation}
where $\mathbf{p}_i \in \mathbb{R}^3$ is the position, $\mathbf{q}_i \in \mathbb{R}^4$ is the orientation quaternion, $\mathbf{s}_i \in \mathbb{R}^3$ is the scale, $c_i \in \{1,\dots,C\}$ is the semantic category label represented as a $C$-dimensional one-hot vector which conditioned as learned soft-embeddings, and $N_o$ is the number of objects in $x_t$.

Each element is mapped to a $d_m$-dimensional token:
\begin{equation}
\mathbf{h}_i = \mathbf{f}_i^{(0)} + \mathbf{e}^{\text{pos}}_i + \mathbf{e}^{\text{dom}}_i,
\end{equation}
where $\mathbf{f}_i^{(0)} = \text{MLP}([\mathbf{p}_i, \mathbf{q}_i, \mathbf{s}_i, \text{Emb}(c_i)])$
encodes geometry and semantics, $\mathbf{e}^{\text{pos}}_i$ is sinusoidal positional encoding~\cite{vaswani2017attention}, and $\mathbf{e}^{\text{dom}}_i$ is a learned indoor/outdoor domain embedding parameterized by a small trainable embedding vector per domain, following domain-adaptive encodings as in~\cite{chen2022vision}.


We predict a tokenizability score $\tau_i \in [0,1]$ for each object at the same timestep:
\begin{equation}
\tau_i = \sigma\left( \mathbf{w}_\tau^\top \, \text{MLP}(\mathbf{f}_i^{(0)}) \right),
\end{equation}


Then, a coarse- or fine-grained route check is performed on tokens based on their tokenizability scores, by comparing them against a learned gating threshold $\gamma$, which determines whether each token is classified as a coarse token $\mathcal{T}_{\text{coarse}}$ or a fine-grained token $\mathcal{T}_{\text{fine}}$.
\begin{equation}
\mathcal{T}_{\text{coarse}} = \{\mathbf{h}_i \mid \tau_i < \gamma \}, \qquad
\mathcal{T}_{\text{fine}} = \{\mathbf{h}_i \mid \tau_i \geq \gamma \}.
\label{eq:coarseFineCheck}
\end{equation}
In our implementation $\gamma$ is a learnable scalar (optimized jointly with network parameters). 

\subsection{Coarse and Fine Feature Extraction}
\label{sec:feature_extraction}

Tokens $\mathcal{T}_{\text{coarse}}$ are passed through a lightweight 3D CNN to extract coarse-level features $F_{\text{coarse}}$, which are essential for constructing exteriors (e.g., large-scale structural elements like buildings or terrain): 
\begin{equation}
F_{\text{coarse}} = \text{CNN}_{\text{coarse}}(\mathcal{T}_{\text{coarse}}), \quad F_{\text{coarse}} \in \mathbb{R}^{N_c \times d_m}.
\label{eq:coarseFeatsExtract}
\end{equation}

Tokens $\mathcal{T}_{\text{fine}}$ are used to construct a spatial adjacency graph $G_{\text{fine}}=(V,E)$: each node corresponds to a fine token and edges connect pairs with Euclidean distance $\le \delta_f$. A two-layer GNN refines these local features:
\begin{equation}
F_{\text{fine}} = \text{GNN}_{\text{fine}}(\mathcal{T}_{\text{fine}}, G_{\text{fine}}), \quad F_{\text{fine}} \in \mathbb{R}^{N_f \times d_m},
\end{equation}
with node updates
\begin{equation}
\mathbf{f}_i^{(l+1)} = \text{MLP}\Big( \mathbf{f}_i^{(l)} \,\Vert\, \sum_{j \in \mathcal{A}_i} \text{MLP}(\mathbf{f}_j^{(l)}) \Big).
\end{equation}
where $\mathcal{A}_i$ denotes the set of neighboring nodes of token $i$ in $G_{\text{fine}}$, and $\Vert$ indicates vector concatenation.  


\subsection{Cross-scale Progressive Attention}
\label{sec:hst}



Cross-scale Progressive Attention is composed of stacked cross-scale attention blocks that alternate between top-down (coarse$\rightarrow$fine) and bottom-up (fine$\rightarrow$coarse) interactions:  
\begin{equation}
\text{Top-down: } 
Q = F_{\text{fine}}, \quad K,V = F_{\text{coarse}},
\end{equation}
\begin{equation}
\text{Bottom-up: } 
Q = F_{\text{coarse}}, \quad K,V = F_{\text{fine}}.
\end{equation}

Each block follows the pattern  
\begin{equation}
F' = \text{FC}\!\left(\text{Attn}(\text{LN}(Q), \text{LN}(K), \text{LN}(V))\right) + F,
\end{equation}
where $F$ is the residual input to the block. By stacking $L$ alternating top-down and bottom-up blocks, coarse tokens propagate structural context while fine tokens inject local detail. The outputs are concatenated and projected to form the final feature:
\begin{equation}
F_{\text{attn}} \in \mathbb{R}^{(N_c+N_f) \times d_m},
\end{equation}
which condition the UNet denoiser at each diffusion step via cross-attention layers in the UNet.

\subsection{Guidance Objectives}
\label{sec:guidance}
We ensure the physical plausibility and interactivity of generated scenes by guiding the conditional scene diffusion process with physics-based guidance functions. Additionally, the extensibility of fine-grained layout placement must align with coarse-scale structures while ensuring consistency in object categories and spatial relationships. This motivation led us to implement three guidance objectives:

\subsubsection{Collision Avoidance Guidance}
Penalizes spatial overlap above a small tolerance $\delta_d$ using 3D IoU~\cite{zhou2019iou}:
\begin{equation}
\mathcal{L}_{\text{col}}(x_t) = \sum_{i \neq j} \max(0, \text{IoU}(B_i, B_j) - \delta_d),
\end{equation}
where each $B_i$ is an oriented 3D bounding box parameterized by its center $\mathbf{p}_i$, orientation quaternion $\mathbf{q}_i$, and scale $\mathbf{s}_i$, with $B_j$ defined analogously.

\subsubsection{Coarse-to-Fine Guidance}
Encourages fine-grained placement to remain consistent with the coarse-scale structural plan:
\begin{equation}
\mathcal{L}_{\text{c2f}}(x_t) = \sum_{o_i \in \text{fine}} \text{dist}\big( \mathbf{p}_i, \mathcal{R}_{\text{coarse}}(o_i) \big),
\end{equation}
where $\mathcal{R}_{\text{coarse}}(o_i)$ is the spatial region assigned by the corresponding coarse token (determined via Euclidean distance to coarse centers), following the idea of region decomposition in~\cite{paschalidou2021atiss}, and $\text{dist}(\cdot,\cdot)$ denotes the Euclidean distance between the object center $\mathbf{p}_i$ and the region's centroid.

\subsubsection{Semantic Constraint Guidance}
Ensures object categories and spatial relations follow learned semantic priors:
\begin{equation}
\mathcal{L}_{\text{sem}}(x_t) = - \sum_{(o_i, o_j)} \log P_{\text{sem}}(c_i, c_j, r_{ij}),
\end{equation}
where $P_{\text{sem}}$ is an MLP estimated from pairwise statistics of the training set and 
$r_{ij}$ denotes the spatial displacement between objects $o_i$ and $o_j$.

\medskip
We define the combined guidance loss as
\begin{equation}
\mathcal{L}_{\text{guide}}(x_t) = 
\mathcal{L}_{\text{col}}(x_t) + 
\mathcal{L}_{\text{c2f}}(x_t) + 
\mathcal{L}_{\text{sem}}(x_t).
\end{equation}

During training, $\mathcal{L}_{\text{guide}}$ enters the total loss with a small weight (Algorithm~\ref{alg:training_concise}). During inference, $\nabla_{x_t}\mathcal{L}_{\text{guide}}$ is normalized and scaled before adjusting $\mu_\theta$ (Algorithm~\ref{alg:inference_concise}). This is depicted in Fig.~\ref{fig:overview} as a feedback arrow from Guidance Objectives to the denoiser output.

\subsection{Training Algorithm}
Training proceeds by sampling a batch of ground-truth layouts and a noise timestep \(t\), forming the noisy input \(x_t\)
. For each \(x_t\), we compute hierarchy tokens and extract multi-scale conditioning via the coarse/fine branches and the Cross-scale Progressive Attention (Sec.~\ref{sec:hst}), then predict \(\epsilon_\theta(x_t,t)\) and minimize the reconstruction loss
\begin{equation}
\mathcal{L}_{\text{rec}}=\|\epsilon-\epsilon_\theta(x_t,t)\|^2.
\end{equation}

In parallel, we compute differentiable guidance losses \(\mathcal{L}_{\text{guide}}\) on object parameters (using smooth surrogates such as soft IoU, signed distance functions, or soft adjacency). The total training loss is
\begin{equation}
\mathcal{L}=\mathcal{L}_{\text{rec}}+\lambda_{\text{guide}} \mathcal{L}_{\text{guide}},
\end{equation}
where $\lambda_{\text{guide}}$ is a scalar value controlling the influence of guidance. Guidance is applied softly (i.e., small $\lambda_{\text{guide}}$ value) during training to stabilize optimization, while at inference (Algorithm~\ref{alg:inference_concise}), it is always enforced with step-size scaling.  


\begin{algorithm}[h]
\caption{Training Procedure}
\label{alg:training_concise}
Initialize parameters $\theta$ (diffusion UNet) and $\psi$ (embedding).

\For{each batch $(x_0)$ in dataset}{
      Sample timestep $t$ and noise $\epsilon$.  
    
      Generate $x_t = \sqrt{\bar{\alpha}_t}x_0 + \sqrt{1 - \bar{\alpha}_t}\epsilon$.
    
      Compute hierarchy tokens for $x_t$, extract coarse/fine features, and fuse with Cross-scale Progressive Attention (Sec.~\ref{sec:hierarchy_embedding},~\ref{sec:hst}).
    
      Predict noise $\epsilon_\theta(x_t,t)$ and compute reconstruction loss $\mathcal{L}_{\text{rec}}$.
    
      Compute differentiable guidance loss $\mathcal{L}_{\text{guide}}$ on object parameters.
    
      Form total loss $\mathcal{L} = \mathcal{L}_{\text{rec}} + \lambda_{\text{guide}}\mathcal{L}_{\text{guide}}$.
    
      Update $\theta,\psi$ by gradient descent on $\mathcal{L}$.
    
}
\end{algorithm}

\subsection{Inference Algorithm}
Inference performs the reverse diffusion starting from \(x_T\sim\mathcal{N}(0,I)\) and iterating \(t=T,\dots,1\). At each step we compute hierarchy conditioning for the current \(x_t\), predict \(\epsilon_\theta(x_t,t)\), and obtain the denoising mean via the epsilon-parameterization
\begin{equation}
\mu_\theta(x_t,t)=\frac{1}{\sqrt{\alpha_t}}\Big(x_t-\frac{\beta_t}{\sqrt{1-\bar\alpha_t}}\epsilon_\theta(x_t,t)\Big).
\end{equation}

We then evaluate the differentiable guidance loss \(\mathcal{L}_{\text{guide}}\) on object parameters, compute its gradient 
\begin{equation}
g = \nabla_{x_t}\mathcal{L}_{\text{guide}}(x_t),
\end{equation}
normalize it \(\tilde g=g/(\|g\|+\varepsilon)\), and scale it by a timestep-dependent step-size
\begin{equation}
\eta_t = \eta_0\sqrt{1-\bar\alpha_t}.
\end{equation}
The denoising mean is adjusted as
\begin{equation}
\mu'_\theta = \mu_\theta - \eta_t \tilde g,
\end{equation}
and the next state is sampled
\begin{equation}
x_{t-1} \sim \mathcal{N}(\mu'_\theta,\Sigma_\theta(x_t,t)).
\end{equation}

After sampling, normalize quaternions $\mathbf{q}_i \leftarrow \mathbf{q}_i / \|\mathbf{q}_i\|$ is applied for each object. Unlike training, guidance is always applied during inference to enforce physical plausibility and semantic consistency.

\begin{algorithm}[h]
\caption{Inference Procedure}
\label{alg:inference_concise}
Initialize $x_T \sim \mathcal{N}(0, I)$.

\For{$t = T$ to $1$}{
      Compute hierarchy tokens for $x_t$, extract coarse/fine features, and fuse with Cross-scale Progressive Attention (Sec.~\ref{sec:hierarchy_embedding},~\ref{sec:hst}).
    
      Predict $\epsilon_\theta(x_t,t)$ and compute the denoising mean $\mu_\theta(x_t,t)$.
    
      Compute guidance gradient $g = \nabla_{x_t}\mathcal{L}_{\text{guide}}(x_t)$, normalize $\tilde g$, and scale with $\eta_t=\eta_0\sqrt{1-\bar\alpha_t}$.
    
      Adjust mean: $\mu'_\theta = \mu_\theta - \eta_t \tilde g$.

      Sample $x_{t-1} \sim \mathcal{N}(\mu'_\theta, \Sigma_\theta(x_t,t))$.
}

Output $x_0$ as synthesized layout.
\end{algorithm}

\section{Scene Synthesis Experiment}
We first compare our scene generation method with recent approaches. In Sec.~\ref{sec:drone-tasks}, we provide the details of our dataset statistics and its application in drone tasks.

\subsection{Implementation, baseline, and evaluation metrics}
\textbf{Implementation Details.}
Our method is implemented in PyTorch and validated on the 3D-FRONT~\cite{fu20213d} and our dataset. 
Training is performed using the Adam optimizer with a learning rate of $2 \times 10^{-4}$, batch size of 64, and a cosine learning rate scheduler. We train for 800 epochs on a single NVIDIA A100 GPU. During training, coarse-to-fine guidance objectives are applied to encourage logical placement of objects across scales, while collision and semantic consistency losses are included to ensure physically plausible and semantically coherent layouts. At inference time, we employ 100 reverse diffusion steps, which provide a balance between synthesis quality and computational efficiency. 

\textbf{Baselines.}
We compare our framework against established baselines for scene layout synthesis. ATISS~\cite{paschalidou2021atiss} employs an autoregressive transformer to generate indoor scenes. Diffusion-SDF~\cite{chou2023diffusion} uses a diffusion model with signed distance fields to model object placements. DiffuScene~\cite{tang2024diffuscene} is a compositional diffusion model that generates scenes without explicit hierarchical modeling. PhyScene~\cite{yang2024physcene} uses physical constraints to generate indoor environments with layouts and articulated objects. 

\textbf{Evaluation Metrics.} We evaluate generated layouts using different metrics: FID~\cite{heusel2017gans} and KID~\cite{binkowski2018demystifying} measure perceptual similarity. Collision Rate (CR) quantifies physical plausibility as the percentage of object pairs with $\text{IoU} > 0.01$. Coarse-to-Fine Consistency (CFC) evaluates hierarchical alignment by computing the average normalized distance between fine placements and their assigned coarse regions. Semantic Plausibility (SP) measures the similarity with spatial category priors, calculated as the negative log-likelihood under empirical pairwise category distributions from the training set. Metrics are averaged over 1,000 scenes per method.

\begin{table}[h]
\centering
\caption{Quantitative results on scene synthesis. 
}
\label{tab:synthesis_results}
\resizebox{\linewidth}{!}{
\begin{tabular}{lccccc}
\toprule
\textbf{Method} & \textbf{FID}$\downarrow$ & \textbf{KID}$\downarrow$ & \textbf{CR}(\%)$\downarrow$ & \textbf{CFC}$\downarrow$ & \textbf{SP}$\downarrow$ \\
\midrule
\rowcolor[HTML]{EFEFEF}\multicolumn{6}{c}{\textbf{Our Dataset}} \\
ATISS~\cite{paschalidou2021atiss}          & 45.2 & 0.032 & 12.5 & 0.21 & 3.8 \\
Diffusion-SDF~\cite{chou2023diffusion}  & 38.7 & 0.028 & 10.1 & 0.18 & 3.5 \\
DiffuScene~\cite{tang2024diffuscene}      & 32.4 & 0.025 & 8.3  & 0.15 & 3.2 \\
PhyScene~\cite{yang2024physcene}       & 29.8 & 0.023 & 7.1  & 0.13 & 3.0 \\
\rowcolor[HTML]{EFEFEF}\textbf{Ours}           & \textbf{27.3} & \textbf{0.021} & \textbf{6.2} & \textbf{0.12} & \textbf{2.7} \\
\midrule
\rowcolor[HTML]{EFEFEF}\multicolumn{6}{c}{\textbf{3D-FRONT Dataset}~\cite{fu20213d}} \\
ATISS~\cite{paschalidou2021atiss}          & 42.1 & 0.030 & 11.8 & 0.19 & 3.6 \\
Diffusion-SDF~\cite{chou2023diffusion}  & 35.6 & 0.026 & 9.4  & 0.16 & 3.3 \\
DiffuScene~\cite{tang2024diffuscene}      & 30.2 & 0.023 & 7.6  & 0.14 & 3.0 \\
PhyScene~\cite{yang2024physcene}       & 27.9 & 0.021 & 6.3  & 0.12 & 2.7 \\
\rowcolor[HTML]{EFEFEF}\textbf{Ours }          & \textbf{25.8} & \textbf{0.019} & \textbf{5.5} & \textbf{0.11} & \textbf{2.5} \\
\bottomrule
\end{tabular}
}
\end{table}
\vspace{-2ex}

\subsection{Scene Generation Results}
\begin{figure}[t]
  \centering
\setlength{\tabcolsep}{3pt}
\begin{tabular}{ccc}
\shortstack{\includegraphics[width=0.48\linewidth]{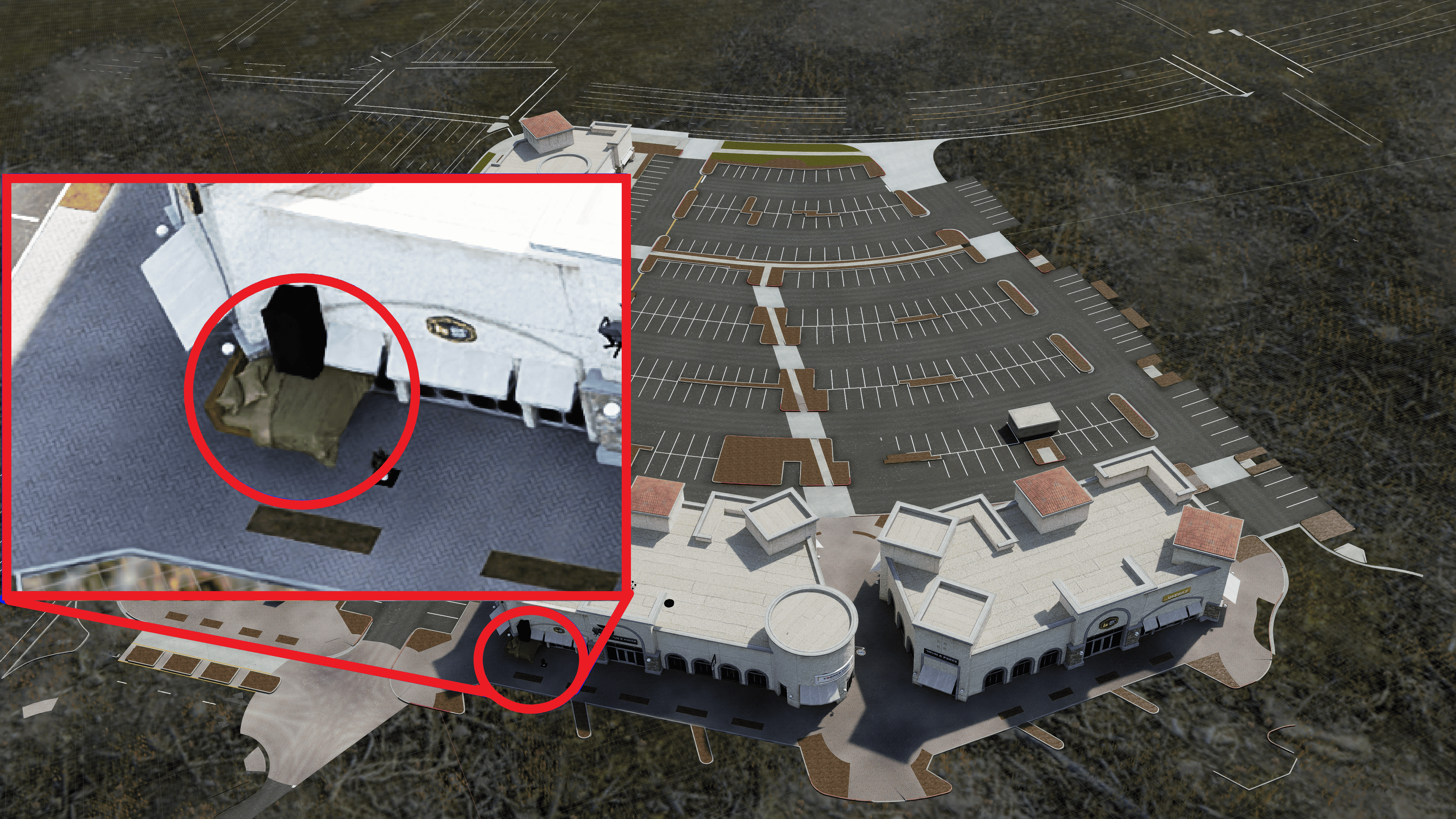}\\\footnotesize (a) SDF~\cite{chou2023diffusion}}&
\shortstack{\includegraphics[width=0.48\linewidth]{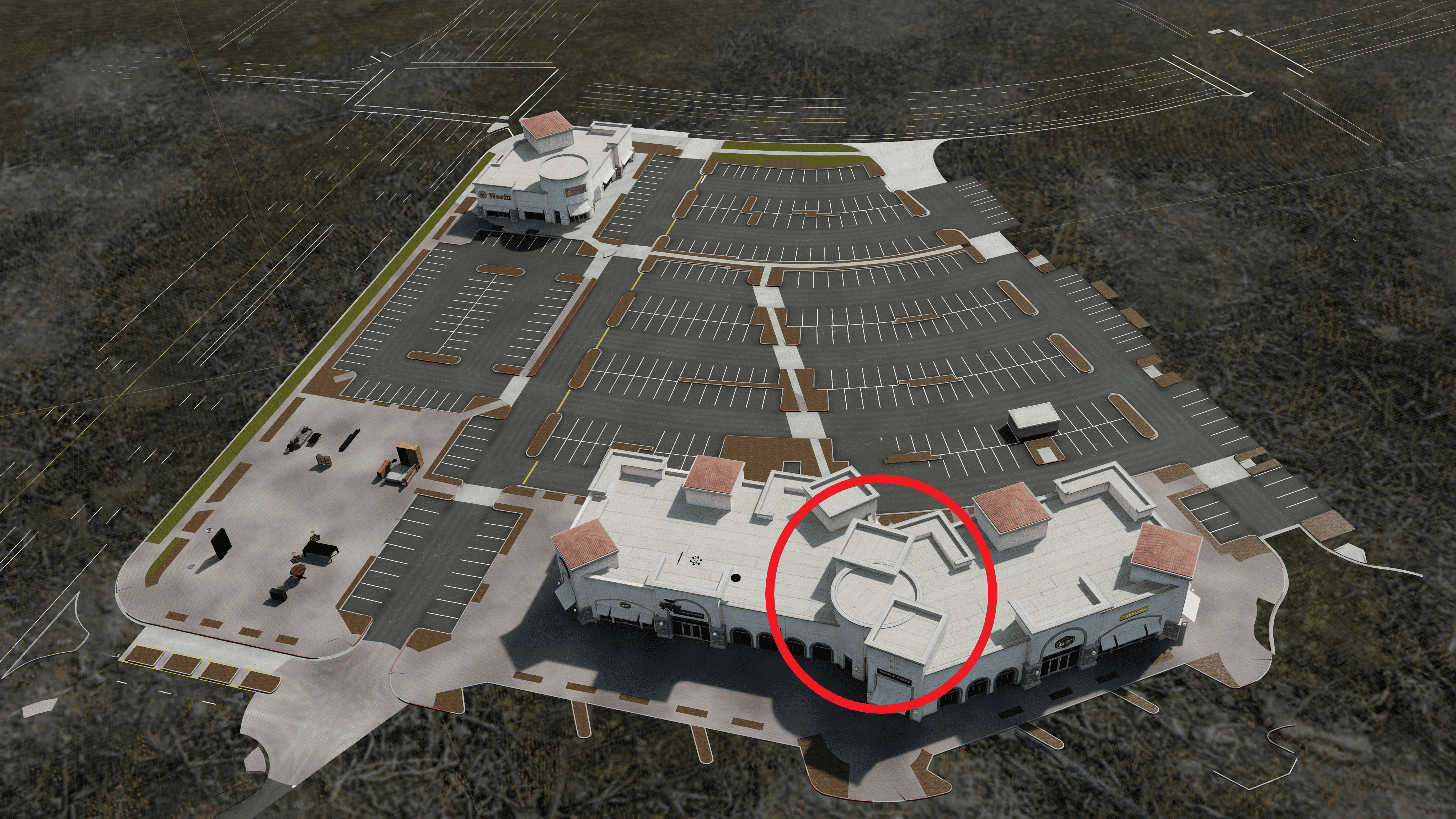}\\ \footnotesize (b) DiffuScene~\cite{tang2024diffuscene}}\\
\shortstack{\includegraphics[width=0.48\linewidth]{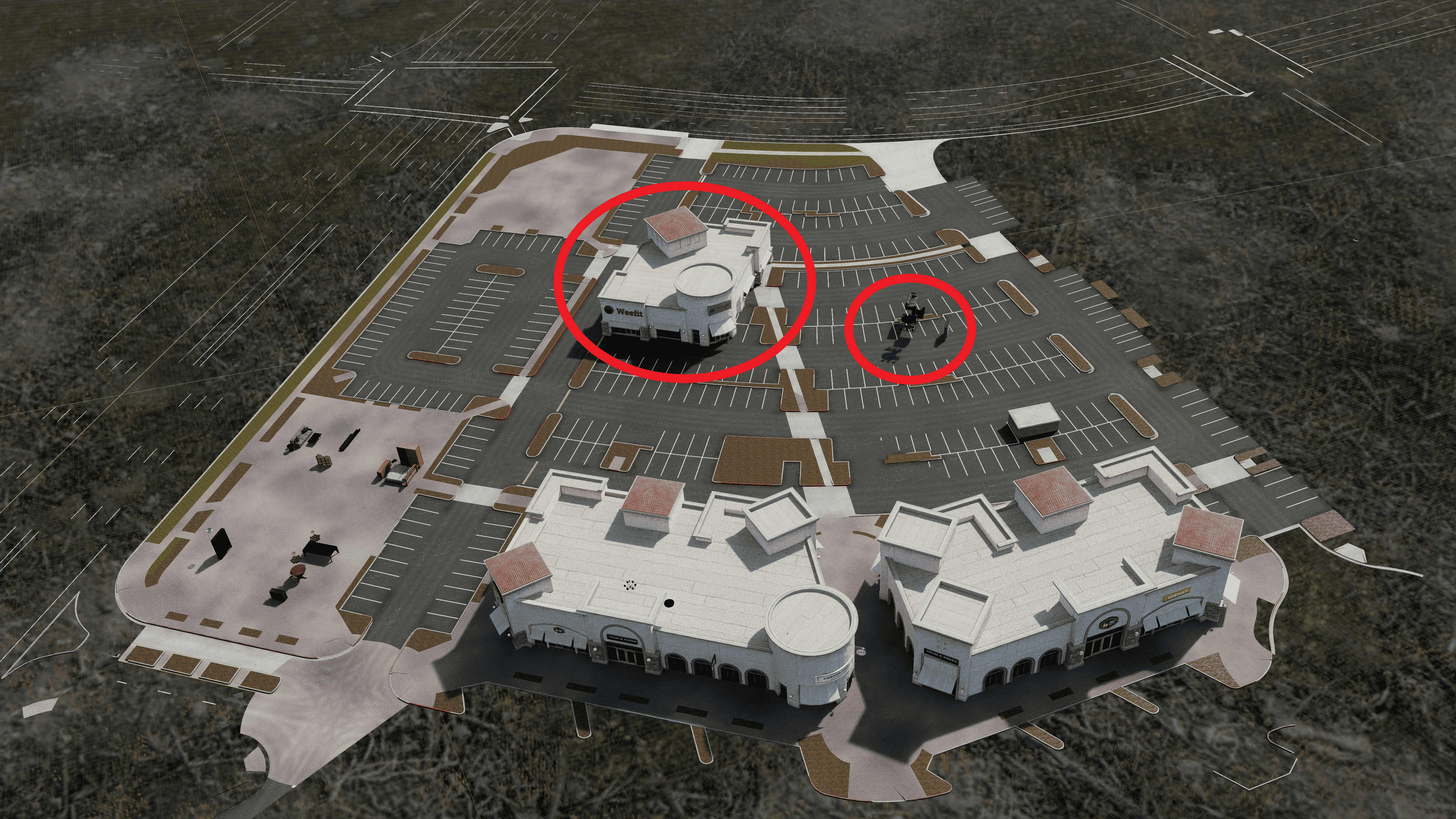}\\\footnotesize (c) PhyScene~\cite{yang2024physcene}}&
\shortstack{\includegraphics[width=0.48\linewidth]{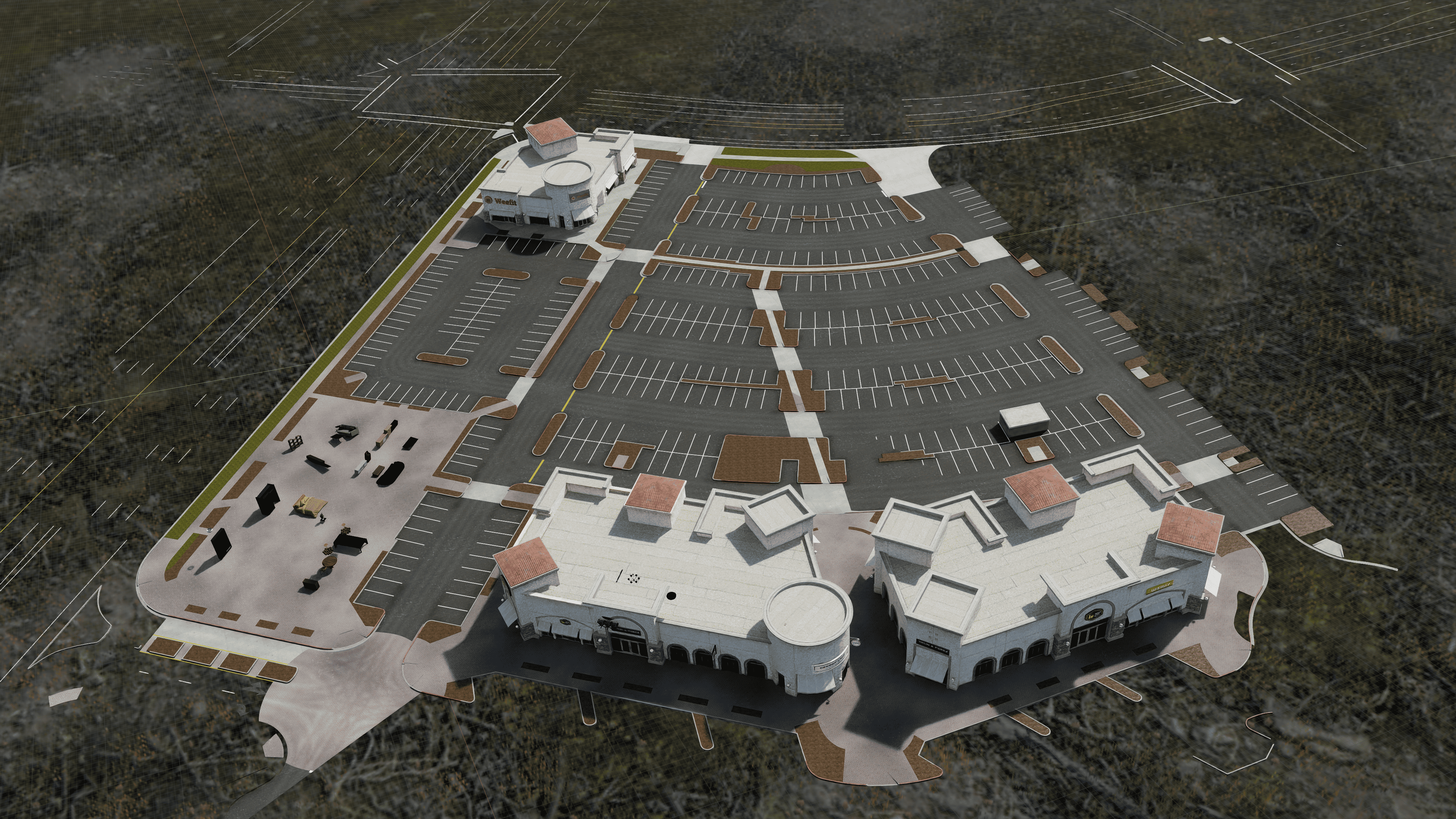}\\ \footnotesize (d) Ours}\\
\end{tabular}
\vspace{-2ex}
    \caption{Outdoor scene generation visual comparison. The red circle shows the collision or incorrect position.  
    }
    \label{fig:outdoorCompare}
\end{figure}

\begin{figure}[t]
  \centering
\setlength{\tabcolsep}{3pt}
\begin{tabular}{ccc}
\shortstack{\includegraphics[width=0.48\linewidth]{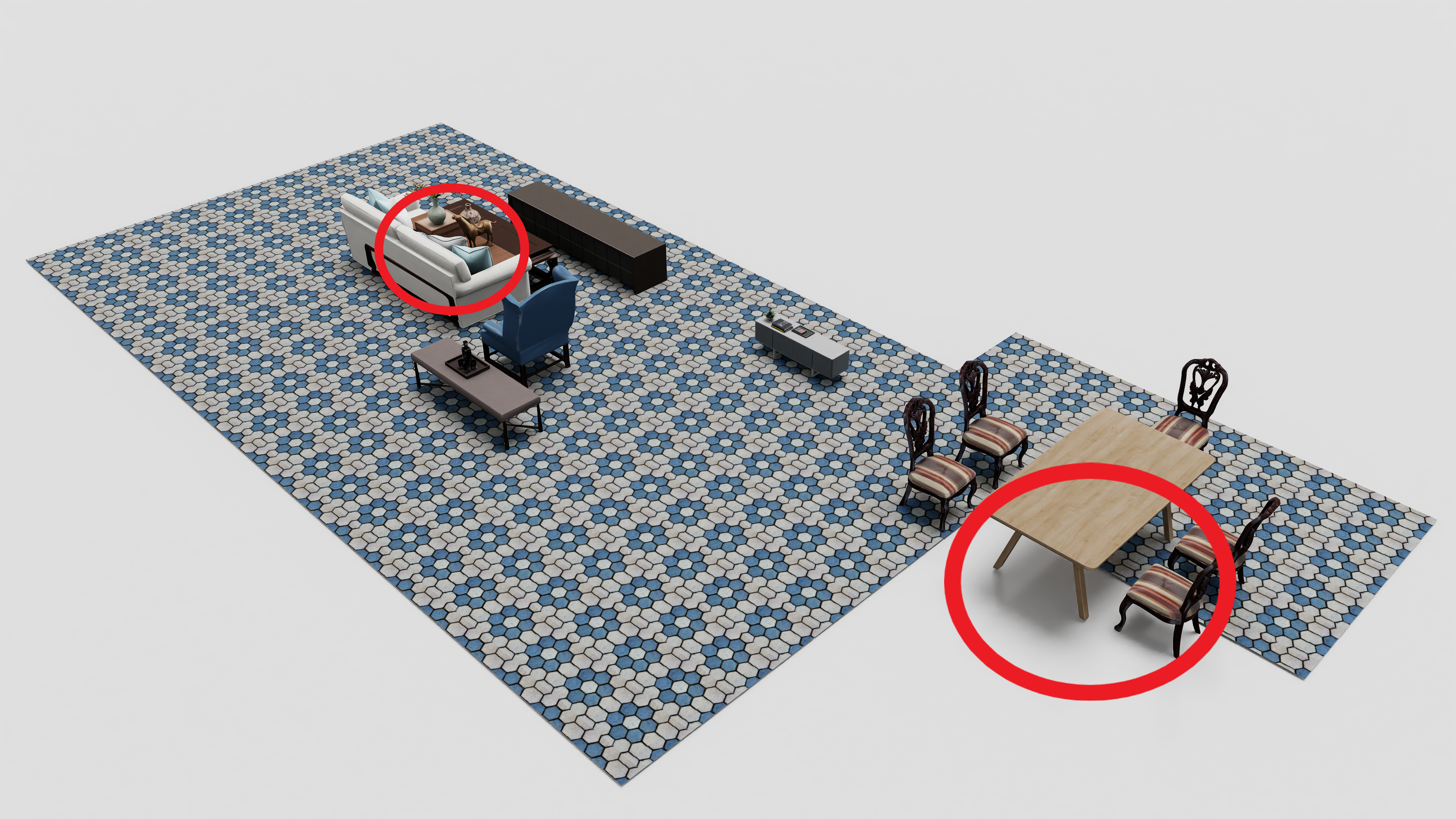}\\\footnotesize (a) SDF~\cite{chou2023diffusion}}&
\shortstack{\includegraphics[width=0.48\linewidth]{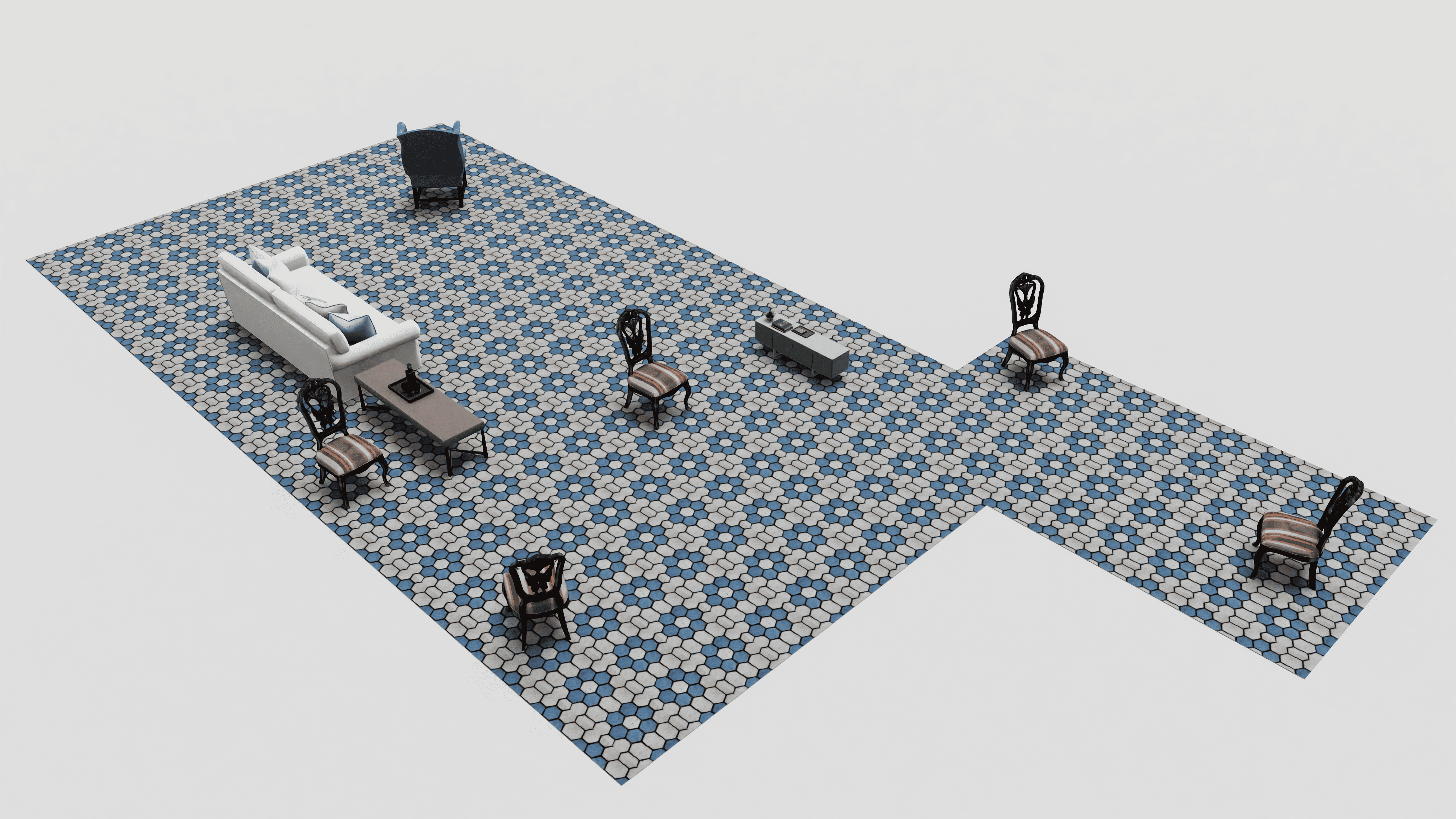}\\ \footnotesize (b) DiffuScene~\cite{tang2024diffuscene}}\\
\shortstack{\includegraphics[width=0.48\linewidth]{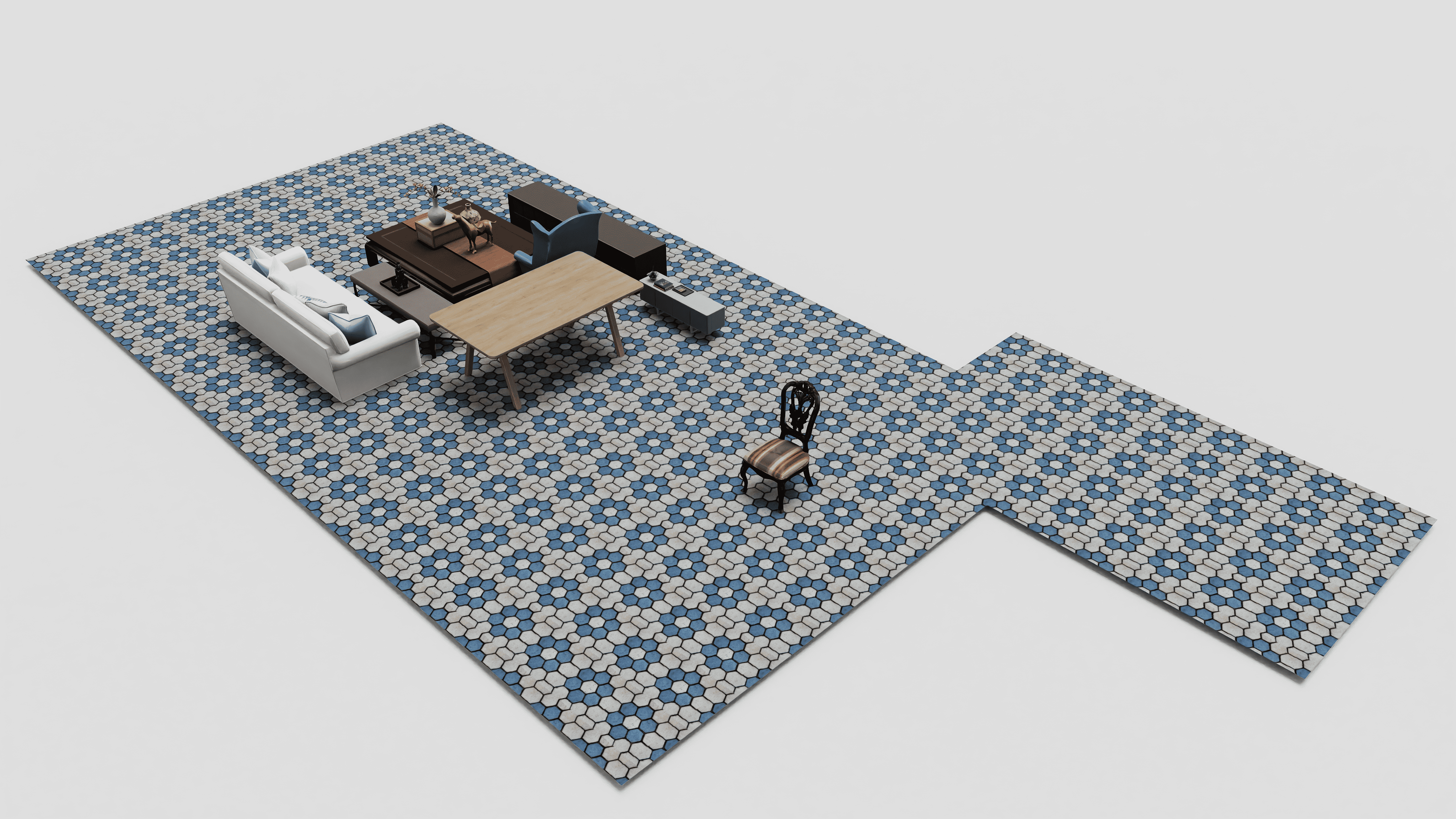}\\\footnotesize (c) PhyScene~\cite{yang2024physcene}}&
\shortstack{\includegraphics[width=0.48\linewidth]{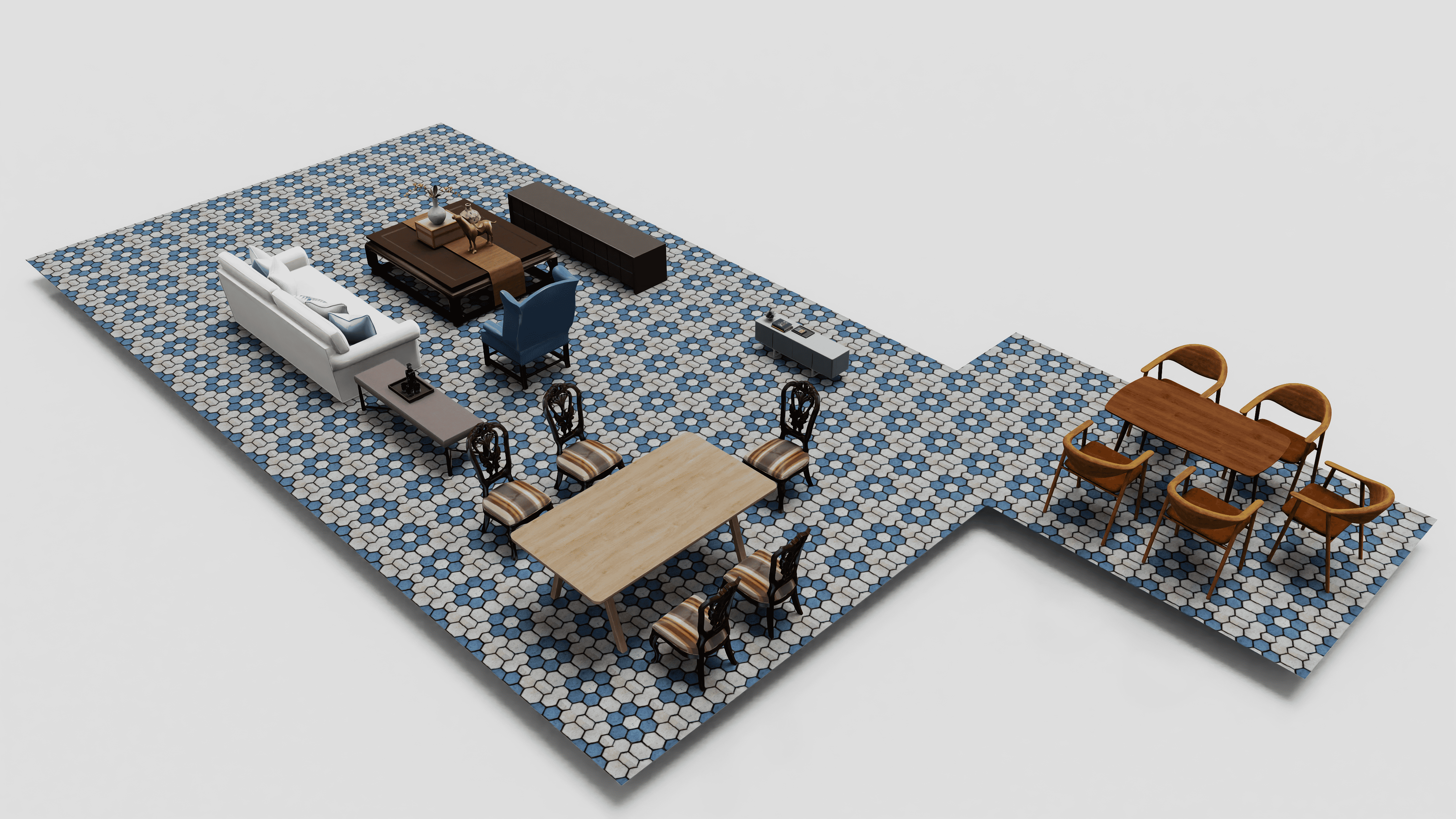}\\ \footnotesize (d) Ours}\\
\end{tabular}
\vspace{-2ex}
    \caption{Indoor scene generation visual comparison. The red circle shows the collision or incorrect position. 
    }
    \label{fig:indoorCompare}
    \vspace{0ex}
\end{figure}

\begin{figure}[t]
  \centering
\setlength{\tabcolsep}{3pt}
\begin{tabular}{cc}
\shortstack{\includegraphics[width=0.48\linewidth]{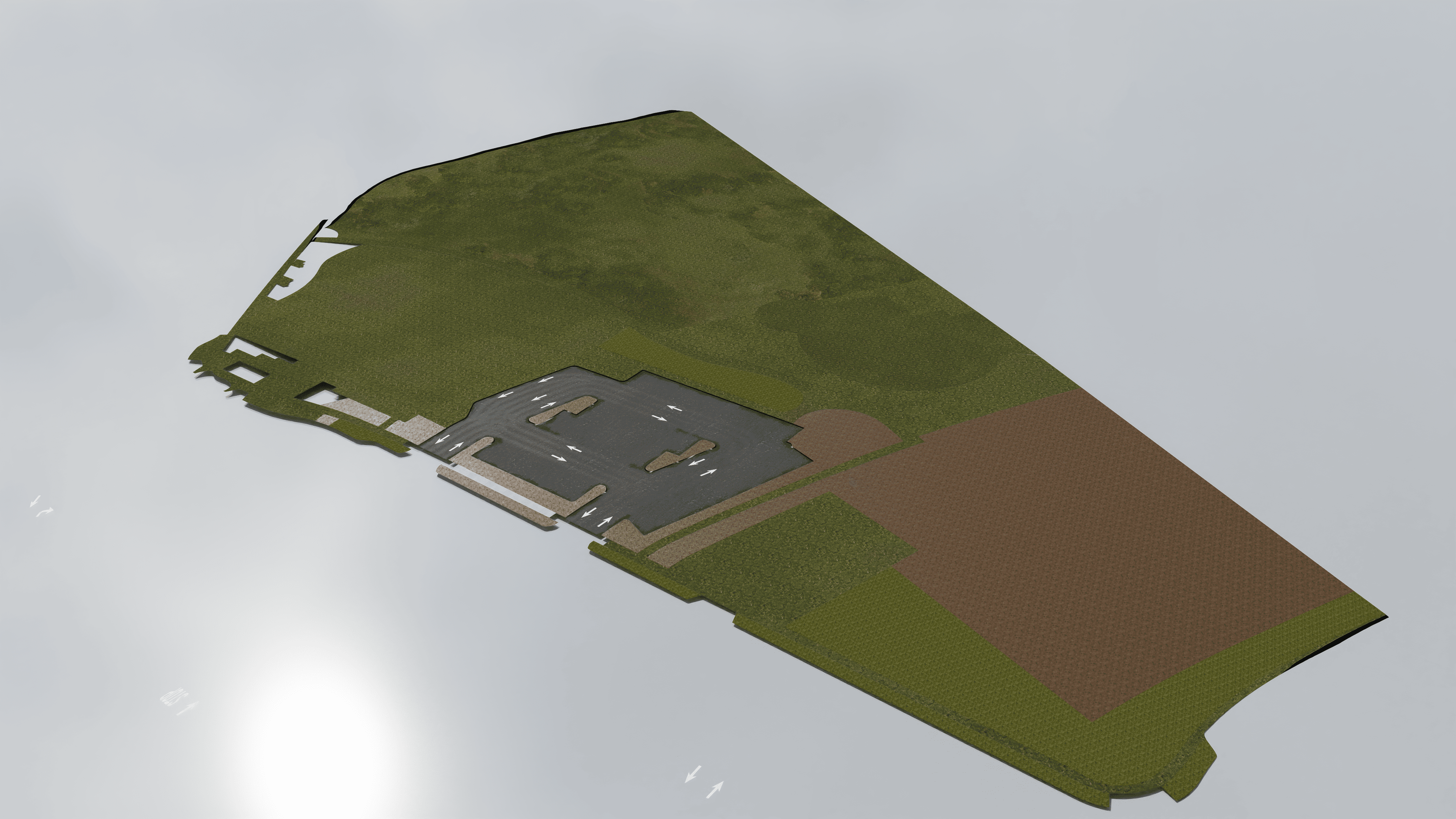}\\\footnotesize (a) Layout}&
\shortstack{\includegraphics[width=0.48\linewidth]{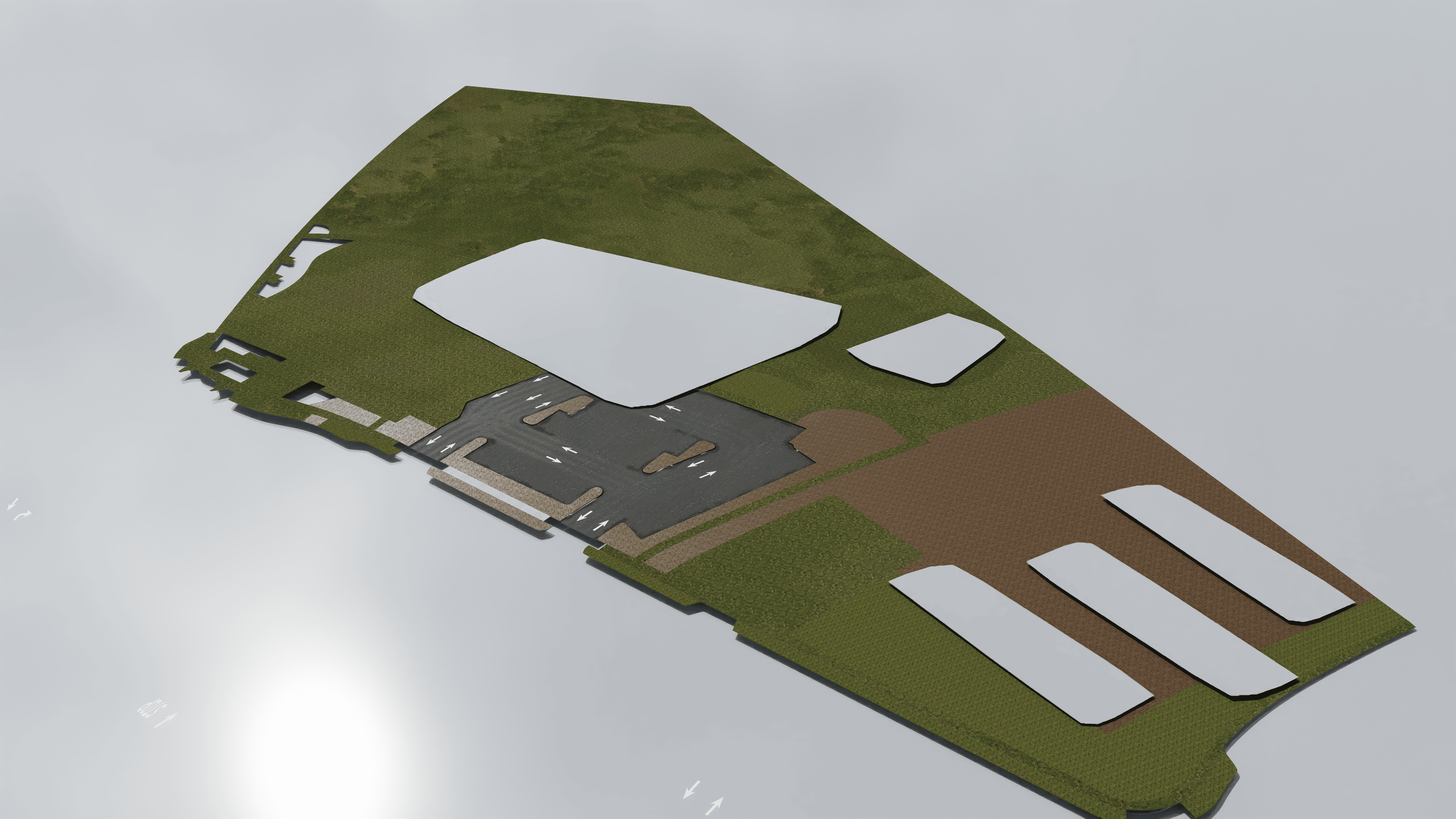}\\ \footnotesize (b) Tokenization}\\
\shortstack{\includegraphics[width=0.48\linewidth]{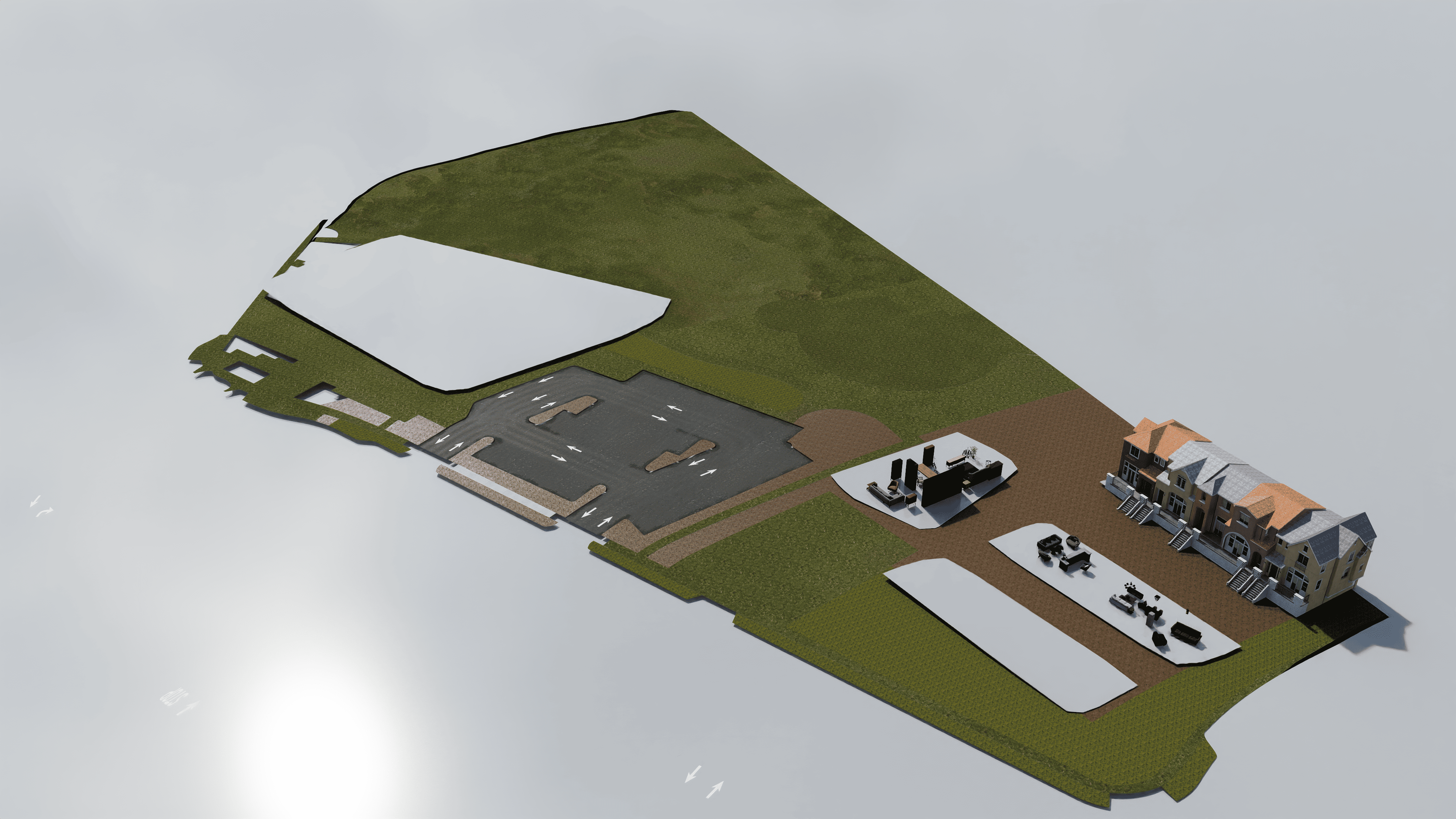}\\\footnotesize (c) Generation w. Constraints}&
\shortstack{\includegraphics[width=0.48\linewidth]{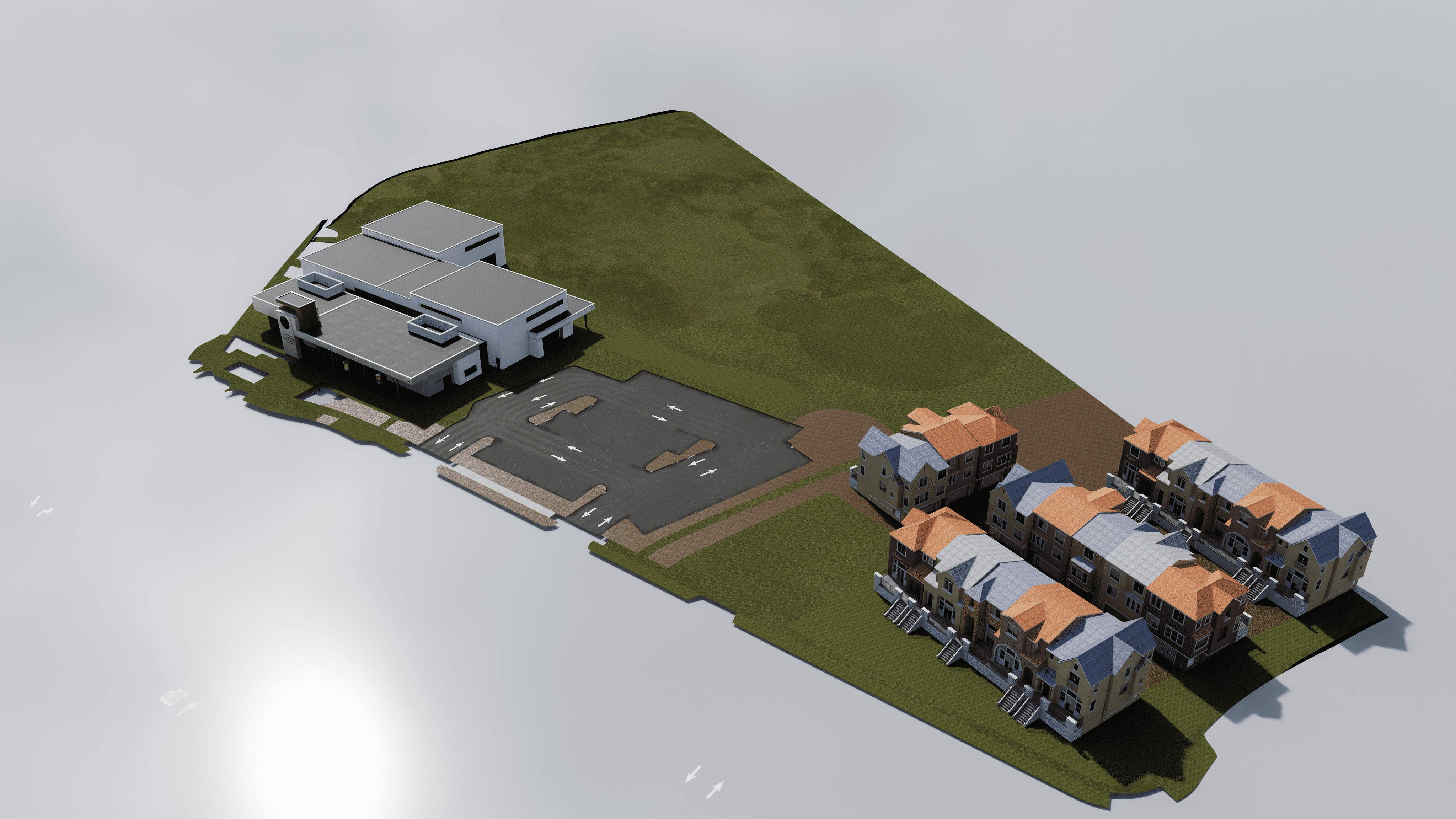}\\ \footnotesize (d) Finalization}\\
\end{tabular}
\vspace{-1ex}
    \caption{The generation sequence of objects in our method.
    }
    \label{fig:SceneGenProgress}
\end{figure}

Table~\ref{tab:synthesis_results} shows quantitative results of our method. This table shows that our method outperforms baselines in all metrics. Qualitatively, Fig.~\ref{fig:outdoorCompare} illustrates that the generated layouts in our model produce coherent hierarchies, grouping fine objects (e.g., tables, chairs, sofas) within coarse structures (e.g., room partitions), whereas baselines often exhibit overlaps or implausible placements. On the 3D-FRONT dataset, similar trends hold on both qualitative and quantitative evaluations. Our method yields more realistic indoor arrangements compared with other solutions (Fig.~\ref{fig:indoorCompare}). We also visualize the generation progress of our guided substitution for diffirent objects in Fig.~\ref{fig:SceneGenProgress}.

\subsection{Ablation Study on Guidance}
To assess the impact of guidance objectives, we conduct ablations by removing individual components during training and inference. Table~\ref{tab:ablation_guidance} shows results on our dataset. Removing Collision Avoidance increases CR by 40\%, indicating its role in physical plausibility. Without Coarse-to-Fine Guidance, CFC degrades by 25\%, leading to misaligned hierarchies. Semantic Constraint Guidance is crucial for SP, with its removal causing a 30\% worsening. Combining all guidance yields the best performance, confirming their complementary nature. Note that our inference time is approximately $2$ minutes per scene on an NVIDIA A100 GPU, comparable to other diffusion baselines.

\begin{table}[h]
\centering
\caption{Ablation on guidance objectives.}
\label{tab:ablation_guidance}
\resizebox{\linewidth}{!}{
\begin{tabular}{lcccc}
\toprule
\textbf{Configuration} & \textbf{FID}$\downarrow$ & \textbf{CR}(\%)$\downarrow$ & \textbf{CFC}$\downarrow$ & \textbf{SP}$\downarrow$ \\
\midrule
\rowcolor[HTML]{EFEFEF}\textbf{Ours (full)}        & \textbf{27.3} & \textbf{6.2} & \textbf{0.12} & \textbf{2.7} \\
w/o Collision      & 32.1 & 8.7 & 0.13 & 2.8 \\
\rowcolor[HTML]{EFEFEF}w/o C2F            & 30.5 & 6.5 & 0.15 & 2.9 \\
w/o Semantic       & 31.8 & 6.4 & 0.13 & 3.5 \\
\rowcolor[HTML]{EFEFEF}w/o All Guidance   & 35.4 & 9.2 & 0.17 & 3.9 \\
\bottomrule
\end{tabular}
}
\end{table}

\section{AeroScene Dataset for Aerial Robotic Tasks}
\label{sec:drone-tasks}

\subsection{AeroScene Dataset Statistic and Labels}
Using our proposed method, we create the AeroScene dataset with more than 1000 scenes, which are specifically curated to emphasize multi-scale hierarchical structures in indoor and outdoor environments. All scenes are embedded into NVIDIA Isaac Sim~\cite{makoviychuk2021isaac} to facilitate downstream drone-related tasks. Scene creation began with initializing empty indoor (e.g., rooms, offices) and outdoor (e.g., parks, urban areas) settings, where base layouts were formed by importing pre-built assets such as walls, floors, and terrain. Objects were then placed in a hierarchical manner: coarse-scale structural elements like buildings, walls, or terrains were positioned first to define the overall structure, followed by fine-scale details such as utensils, decorations, or debris, which were arranged relative to the larger elements to preserve logical groupings and spatial relationships. Finally, all objects were annotated with bounding boxes, orientations, scales, and semantic categories, while hierarchies explicitly linked fine objects to their coarse parents. In addition, interaction areas suitable for drone landing were annotated to support downstream robotics tasks. Specifically, all scenes are stored within a consistent schema: each object has \texttt{id}, \texttt{scene\_id}, \texttt{parent\_id}, \texttt{category\_id}\,$\in\{1,\dots,C\}$, \texttt{bbox} given as $(\mathbf{p},\mathbf{q},\mathbf{s})$ where $\mathbf{p}\in\mathbb{R}^3$ (center), $\mathbf{q}\in\mathbb{R}^4$ (quaternion), $\mathbf{s}\in\mathbb{R}^3$ (local extents), plus precomputed \texttt{bbox\_corners} for fast IoU tests, a \texttt{domain} flag (indoor/outdoor), and optional \texttt{attributes}; interaction areas (e.g., landing zones) are stored as polygonal regions with centroid and radius. The dataset statistical information can be found in Table~\ref{tab:dataset_stats}.


\begin{table}[ht]
\centering
\caption{Statistics of our dataset. \vspace{-2ex}}
\label{tab:dataset_stats}
\resizebox{\linewidth}{!}{
\begin{tabular}{l|ccc}
\hline
\rowcolor[HTML]{EFEFEF}\textbf{Criteria} & \multicolumn{1}{c|}{\textbf{Train}} & \multicolumn{1}{c|}{\textbf{Test}} & \textbf{Total} \\ \hline
\textit{\#Scenes} & \multicolumn{1}{c|}{812} & \multicolumn{1}{c|}{204} & 1016 \\ \hline
\rowcolor[HTML]{EFEFEF}\textit{\#Objects} & \multicolumn{1}{c|}{122,356} & \multicolumn{1}{c|}{37,654} & 160,010 \\ \hline
\textit{Avg. objects/scene} & \multicolumn{1}{c|}{149} & \multicolumn{1}{c|}{152} & 149 \\ \hline
\quad \textit{Avg. objs / coarse-scale level} & \cellcolor[HTML]{EFEFEF}35 & \cellcolor[HTML]{EFEFEF}32 & \cellcolor[HTML]{EFEFEF}34 \\ 
\quad \textit{Avg. objs / fine-scale level} & 111 & 114 & 112 \\ \hline
\rowcolor[HTML]{EFEFEF}\textit{Avg. Landing Areas for Drones} & \multicolumn{1}{c|}{55} & \multicolumn{1}{c|}{53} & 54 \\ \hline
\quad Small Drone (Avg.)  & 42 & 43 & 42 \\ 
\quad Medium Drone(Avg.) & \cellcolor[HTML]{EFEFEF}11 & \cellcolor[HTML]{EFEFEF}13  & \cellcolor[HTML]{EFEFEF}12 \\ 
\quad Large Drone(Avg.)  & 4  & 4  & 5  \\ \hline
\rowcolor[HTML]{EFEFEF}\textit{\#Categories (coarse-scale)} & \multicolumn{3}{c}{23} \\ \hline
\textit{\#Categories (fine-scale)} & \multicolumn{3}{c}{47} \\ \hline
\end{tabular}
}
\end{table}

\subsection{AeroScene for Navigation and Interaction Tasks}

To demonstrate the practical use of the AeroScene dataset, we define a unified aerial robotics task that combines \emph{long-range navigation} and \emph{close-range physical interaction} within a single scene. In this setup, an aerial robot starts at a designated location and must autonomously navigate to a \emph{pre-annotated interaction area} before performing a controlled landing or perching maneuver. This task demonstrates how AeroScene’s realistic and richly annotated environments can evaluate both high-level planning and fine-grained physical interaction capabilities under diverse conditions.

\textbf{Task Design.}  The drone task is divided into two sequential phases:  \textit{(i) Navigation Phase:} The drone uses global semantic information and dynamics constraints to plan a trajectory to the target area \cite{mueller2015computationally}. A geometric controller \cite{lee2010geometric} executes this path, ensuring stable navigation through complex environments. \textit{(ii) Interaction Phase:} Once near the target, the drone switches to local sensing. Point-cloud data is analyzed to identify surface normals, slope, and clearance, allowing the system to select a safe landing or perching zone. The drone then performs a precise descent and touchdown.

\begin{figure}[t]
  \centering
\includegraphics[width=0.95\linewidth]{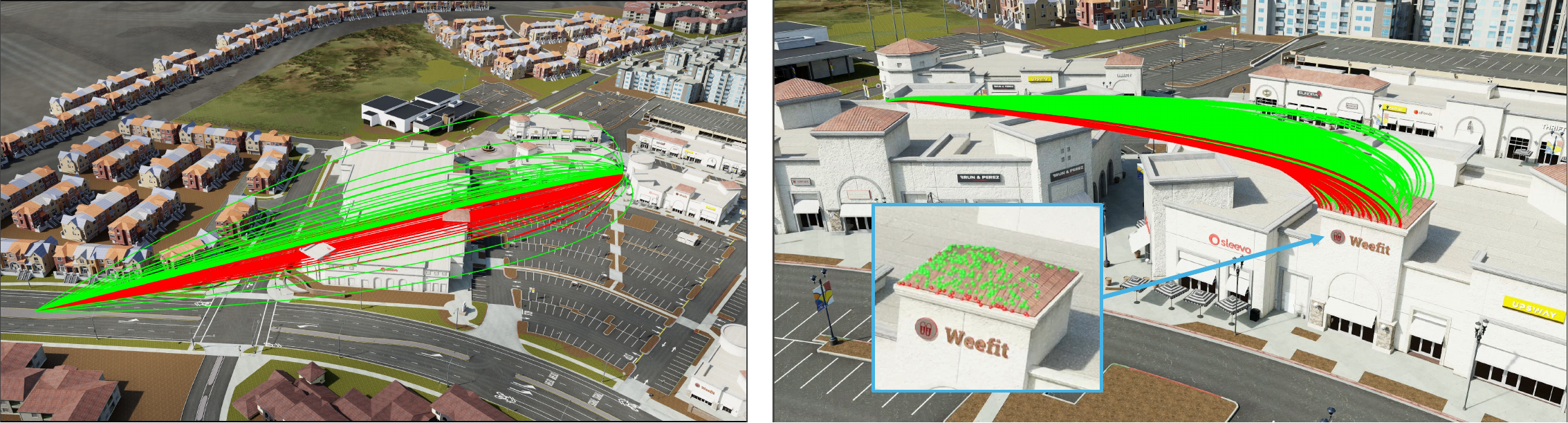}
\vspace{-1ex}
    \caption{Generated navigation and interaction trajectories for the example mission: landing on the red roof of the Weefit building. Green lines represent feasible navigation and interaction trajectories, and red lines denote failed attempts. Sampled point clouds are displayed within the blue box, with red dots indicating failure landing points.
    \vspace{-4ex}}
    \label{fig:GeneratedTrajectories}
\end{figure}

\textbf{Example Scenario.} Fig.~\ref{fig:GeneratedTrajectories} illustrates a representative mission: navigating a generated urban-style scene to land on the red roof of the Weefit building in the scene. In this example, the environment contains detailed objects and varying elevations, requiring the drone to plan a path and then transition to close-proximity perception for accurate landing. This scenario, combined with AeroScene’s dataset generation pipeline, demonstrates the usefulness of our work in benchmarking aerial robotics algorithms in perception, mapping, trajectory planning, and interaction control.

\textbf{Scene Utility and Data Generation.} 
AeroScene’s hierarchical object labeling, annotated landing zones, and physics-aware surfaces create challenging, realistic environments for aerial autonomy and manipulation research. Each trial not only evaluates planning and control strategies but also generates a rich dataset for future development. For every trajectory, we record:
\begin{itemize}
    \item \textbf{Visual Data:} RGB and depth streams from simulated onboard cameras.
    \item \textbf{State Estimates:} Ground-truth absolute position, orientation, and velocity.
    \item \textbf{Inertial Measurements:} IMU sensor readings for accelerations and angular rates.
    \item \textbf{Control Signals:} Low-level motor or actuator commands used during trajectory execution.
    \item \textbf{Planned and Executed Trajectories:} Waypoints and actual flight paths for benchmarking performance.
\end{itemize}
In total, we record each scene 300 planned trajectories for each scene. Small- and medium-sized drone platforms, i.e., 3DR Iris and AscTec Hummingbird, were used to test the unified navigation and interaction pipeline. Across all trials, the system achieved an overall success rate of 91\%, demonstrating the utility of AeroScene as a challenging yet tractable benchmark for aerial robotics research. Details are summarized in Table~\ref{tab:evaluation_summary}.

\begin{table}[t]
\centering

\caption{
Evaluation summary of navigation and interaction tasks on the AeroScene dataset.
\vspace{-1ex}}
\label{tab:evaluation_summary}
\begin{tabular}{lc}
\hline
\rowcolor[HTML]{EFEFEF}\textbf{Metric} & \textbf{Value} \\ \hline
Drone Platforms & 2  \\
Trajectories per Scene & 300 \\

Overall Success Rate & \textbf{91\%} \\ \hline
\end{tabular}
\vspace{-1ex}
\end{table}

\section{Discussion and Conclusion}
\textbf{Limitations.} The proposed AeroScene, while effective in demonstrating hierarchical scene synthesis, has certain limitations. Current experiments are mainly conducted in simulation, which may not fully represent the diversity and complexity of real-world aerial environments (such as wind conditions). In addition, the framework focuses on static scene layouts, without explicitly modeling temporal dynamics or handling uncertainty from dynamic environments. Therefore, an interesting future work is to extend the synthesis process to generate dynamic elements. Furthermore, performing sim-to-real validation on real aerial robots would be an interesting direction for future work.

\textbf{Conclusion.} We introduced AeroScene, a hierarchical diffusion framework for 3D scene synthesis that combines hierarchy-scale tokenization, multi-branch feature extraction, and a cross-scale attention with gradient-based guidance objectives. Our approach enables structured reasoning across global layouts and local details while enforcing physical and semantic plausibility. Using our method, we generate a large-scale benchmark of 3D environments for drone interaction. 
Our code and dataset are publicly available at \href{https://aioz-ai.github.io/AeroScene/}{aioz-ai.github.io/AeroScene/}.



\bibliographystyle{class/IEEEtran}
\bibliography{class/reference}

\end{document}